\documentclass[10pt,letterpaper,twocolumn,english,british]{article}
\usepackage[T1]{fontenc}
\usepackage[latin9]{inputenc}
\usepackage{color}
\usepackage{array}
\usepackage{units}
\usepackage{multirow}
\usepackage{amsmath}
\usepackage{amssymb}
\usepackage{graphicx}

\makeatletter

\pdfpageheight\paperheight
\pdfpagewidth\paperwidth

\providecommand{\tabularnewline}{\\}

\usepackage{cvpr}
\usepackage{times}
\usepackage{epsfig}
\usepackage{graphicx}

\usepackage{makecell, pict2e}

\usepackage[pagebackref=true,breaklinks=true,letterpaper=true,colorlinks,bookmarks=false]{hyperref}

\cvprfinalcopy 

\def\cvprPaperID{165} 

\ifcvprfinal\fi

\@ifundefined{showcaptionsetup}{}{%
 \PassOptionsToPackage{caption=false}{subfig}}
\usepackage{subfig}
\makeatother

\usepackage{babel}
\begin{document}

\title{Filtered Channel Features for Pedestrian Detection}

\author{Shanshan Zhang \and Rodrigo Benenson\vspace{0.5em}
\\
\begin{tabular}{c}
Max Planck Institute for Informatics\tabularnewline
Saarbrücken, Germany\tabularnewline
\texttt{\small{}firstname.lastname@mpi-inf.mpg.de}\tabularnewline
\end{tabular} \vspace{-0.5em}
\and Bernt Schiele}
\maketitle
\begin{abstract}
This paper starts from the observation that multiple top performing
pedestrian detectors can be modelled by using an intermediate layer
filtering low-level features in combination with a boosted decision
forest. Based on this observation we propose a unifying framework
and experimentally explore different filter families. We report extensive
results enabling a systematic analysis.

Using filtered channel features we obtain top performance on the challenging
Caltech and KITTI datasets, while using only HOG+LUV as low-level
features. When adding optical flow features we further improve detection
quality and report the best known results on the Caltech dataset,
reaching 93\% recall at 1 FPPI. 
\end{abstract}
\makeatletter 
\renewcommand{\paragraph}{%
\@startsection{paragraph}{4}%
{\z@}{0.75ex \@plus 1ex \@minus .2ex}{-1em}%
{\normalfont \normalsize \bfseries}%
}
\makeatother

\section{\label{sec:Introduction}Introduction}

Pedestrian detection is an active research area, with 1000+ papers
published in the last decade%
\footnote{Papers from 2004 to 2014 with \textquotedbl{}pedestrian detection\textquotedbl{}
in the title, according to Google Scholar.%
}, and well established benchmark datasets \cite{Dollar2011Pami,Geiger2012CVPR}.
It is considered a canonical case of object detection, and has served
as playground to explore ideas that might be effective for generic
object detection.

Although many different ideas have been explored, and detection quality
has been steadily improving \cite{Benenson2014Eccvw}, arguably it
is still unclear what are the key ingredients for good pedestrian
detection; e.g. it remains unclear how effective parts, components,
and features learning are for this task.

\begin{figure}
\begin{centering}
\vspace{-0.5em}
\includegraphics[width=1.05\columnwidth]{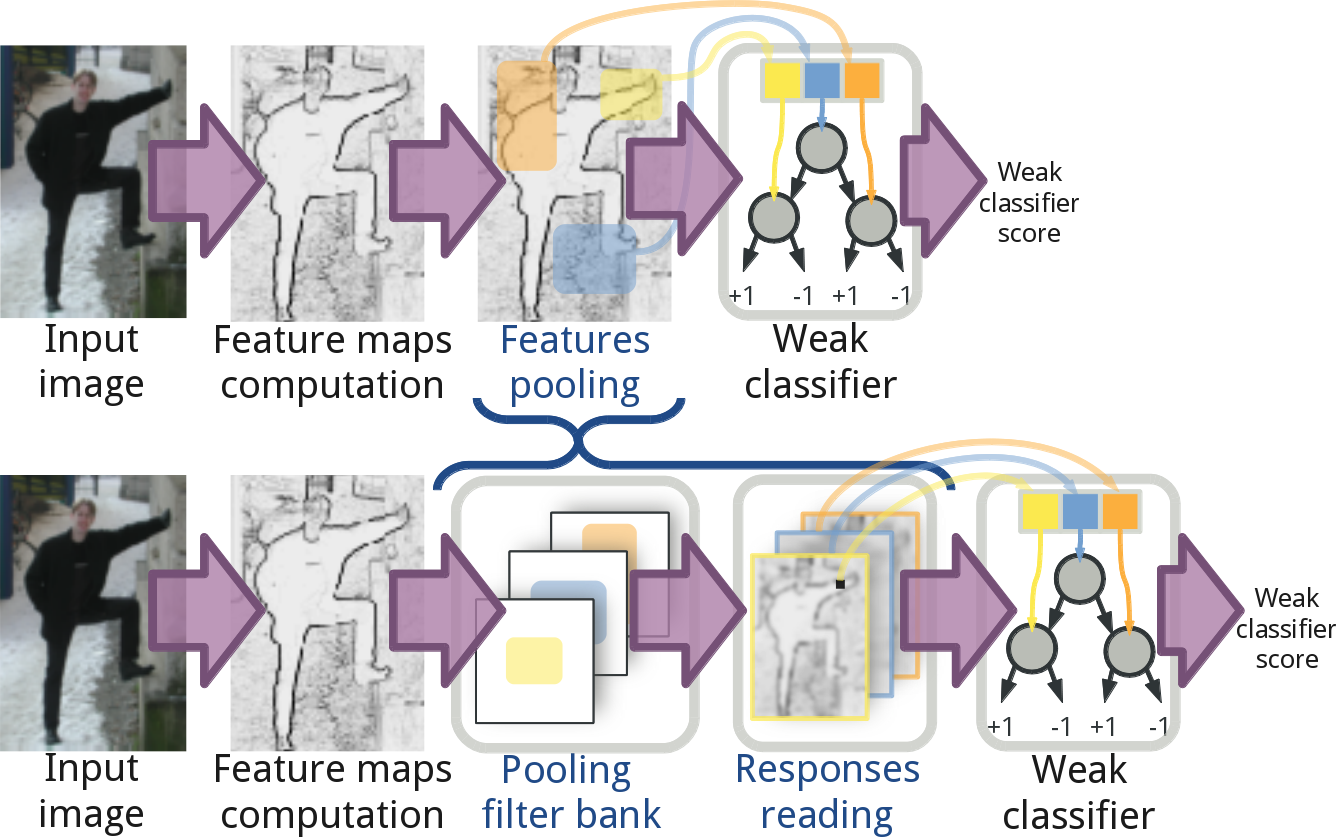}
\par\end{centering}

\protect\caption{\label{fig:Filtered-integral-channels-ilustration}Filtered feature
channels illustration, for a single weak classifier reading over a
single feature channel.\protect \\
Integral channel features detectors pool features via sums over rectangular
regions \cite{Dollar2009Bmvc,Benenson2013Cvpr}. We can equivalently
re-write this operation as convolution with a filter bank followed
by single pixel reads (see \S\ref{sec:Filtered-integral-channels}).
We aim to answer:\emph{}\protect \\
\emph{What is the effect of selecting different filter banks?}}
\vspace{-0.5em}
\end{figure}
Current top performing pedestrian detection methods all point to an
intermediate layer (such as max-pooling or filtering) between the
low-level feature maps and the classification layer \cite{Wang2013IccvRegionlets,Zhang2014CvprInformedHaar,Paisitkriangkrai2014Eccv,Nam2014Nips}.
In this paper we explore the simplest of such intermediary: a linear
transformation implemented as convolution with a filter bank. We propose
a framework for filtered channel features (see figure \ref{fig:Filtered-integral-channels-ilustration})
that unifies multiple top performing methods \cite{Dollar2009Bmvc,Benenson2013Cvpr,Zhang2014CvprInformedHaar,Nam2014Nips},
and that enables a systematic exploration of different filter banks.
With our experiments we show that, with the proper filter bank, filtered
channel features reach top detection quality.

It has been shown that using extra information at test time (such
as context, stereo images, optical flow, etc.) can boost detection
quality. In this paper we focus on the ``core'' sliding window algorithm
using solely HOG+LUV features (i.e. oriented gradient magnitude and
colour features). We consider context information and optical flow
as add-ons, included in the experiments section for the sake of completeness
and comparison with existing methods. Using only HOG+LUV features
we already reach top performance on the challenging Caltech and KITTI
datasets, matching results using optical flow and significantly more
features (such as LBP and covariance \cite{Wang2013IccvRegionlets,Paisitkriangkrai2014Eccv}).

\subsection{\label{sub:Related-work}Related work}

Recent survey papers discuss the diverse set of ideas explored for
pedestrian detection \cite{Enzweiler2009PAMI,Geronimo2010Pami,Dollar2011Pami,Benenson2014Eccvw}.
The most recent survey \cite{Benenson2014Eccvw} indicates that the
classifier choice (e.g. linear/non-linear SVM versus decision forest)
is not a clear differentiator regarding quality; rather the features
used seem more important.

Creativity regarding different types of features has not been lacking.
\textbf{HOG)} The classic HOG descriptor is based on local image differences
(plus pooling and normalization steps), and has been used directly
\cite{Dalal2005Cvpr}, as input for a deformable parts model \cite{Felzenszwalb2010Pami},
or as features to be boosted \cite{Laptev2009Ivc,Nam2011IccvWorkshop}.
The integral channel features detector \cite{Dollar2009Bmvc,Benenson2013Cvpr}
uses a simpler HOG variant with sum pooling and no normalizations.
Many extensions of HOG have been proposed (e.g. \cite{Hou2007Accv,Felzenszwalb2010Pami,Dang2011Kse,Satpathy2014}).
\textbf{LBP)} Instead of using the magnitude of local pixel differences,
LBP uses the difference sign only as signal \cite{Wang2009Iccv,Wang2013IccvRegionlets,Paisitkriangkrai2014Eccv}.
\textbf{Colour)} Although the appearance of pedestrians is diverse,
the background and skin areas do exhibit a colour bias. Colour has
shown to be an effective feature for pedestrian detection and hence
multiple colour spaces have been explored (both hand-crafted and learned)
\cite{Dollar2009Bmvc,Khan2012Cvpr,Khan2013Cvpr,Mathias2014Eccv}.
\textbf{Local structure)} Instead of simple pixel values, some approaches
try to encode a larger local structure based on colour similarities
(soft-cue) \cite{Walk2010Cvpr,Goto2013}, segmentation methods (hard-decision)
\cite{Ott2009Cvpr,Ramanan2007Cvpr,Socarras2012Acivs}, or by estimating
local boundaries \cite{Lim2013Cvpr}. \textbf{Covariance)} Another
popular way to encode richer information is to compute the covariance
amongst features (commonly colour, gradient, and oriented gradient)
\cite{Tuzel2008Pami,Paisitkriangkrai2014Eccv}. \textbf{Etc.)} Other
features include bag-of-words over colour, HOG, or LBP features \cite{Costea2014CVPR};
learning sparse dictionary encoders \cite{Ren2013Cvpr}; and training
features via a convolutional neural network \cite{Sermanet2013Cvpr}.
Additional features specific for stereo depth or optical flow have
been proposed, however we consider these beyond the focus of this
paper. For our flow experiments we will use difference of frames from
weakly stabilized videos (\texttt{SDt}) \cite{Park2013Cvpr}.

All the feature types listed above can be used in the integral channel
features detector framework \cite{Dollar2009Bmvc}. This family of
detectors is an extension of the old ideas from Viola\&Jones \cite{Viola2005IJCV}.
Sums of rectangular regions are used as input to decision trees trained
via Adaboost. Both the regions to pool from and the thresholds in
the decision trees are selected during training. The crucial difference
from the pioneer work \cite{Viola2005IJCV} is that the sums are done
over feature channels other than simple image luminance.

Current top performing pedestrian detection methods (dominating INRIA
\cite{Dalal2005Cvpr}, Caltech \cite{Dollar2011Pami} and KITTI datasets
\cite{Geiger2012CVPR}) are all extensions of the basic integral channel
features detector (named \texttt{ChnFtrs} in \cite{Dollar2009Bmvc},
which uses only HOG+LUV features). \texttt{SquaresChnFtrs} \cite{Benenson2014Eccvw},
\texttt{InformedHaar} \cite{Zhang2014CvprInformedHaar}, and \texttt{LDCF}
\cite{Nam2014Nips}, are discussed in detail in section \ref{sub:Baselines}.
\texttt{Katamari} exploits context and optical flow for improved performance.
\texttt{SpatialPooling(+)} \cite{Paisitkriangkrai2014Eccv} adds max-pooling
on top of sum-pooling, and uses additional features such as covariance,
LBP, and optical flow. Similarly, \texttt{Regionlets} \cite{Wang2013IccvRegionlets}
also uses extended features and max-pooling, together with stronger
weak classifiers and training a cascade of classifiers. Out of these,
\texttt{Regionlets} is the only method that has also shown good performance
on general classes datasets such as Pascal VOC and ImageNet.

In this paper we will show that vanilla HOG+LUV features have not
yet saturated, and that, when properly used, they can reach top performance
for pedestrian detection.

\subsection{\label{sub:Contributions}Contributions}

\begin{itemize}
\item We point out the link between \texttt{ACF} \cite{Dollar2014Pami},
\texttt{(Squares)ChnFtrs} \cite{Dollar2009Bmvc,Benenson2013Cvpr,Benenson2014Eccvw},
\texttt{InformedHaar} \cite{Zhang2014CvprInformedHaar}, and \texttt{LDCF}
\cite{Nam2014Nips}. See section \ref{sec:Filtered-integral-channels}.
\item We provide extensive experiments to enable a systematic analysis of
the filtered integral channels, covering aspects not explored by related
work. We report the summary of $65+$ trained models (corresponding
$\sim\negthickspace10$ days of single machine computation). See
sections \ref{sec:How-many-filters}, \ref{sec:Additional-training-data}
and \ref{sec:Test-set-results}.
\item We show that top detection performance can be reached on Caltech and
KITTI using HOG+LUV features only. We additionally report the best
known results on Caltech. See section \ref{sec:Test-set-results}.
\end{itemize}

\section{\label{sec:Filtered-integral-channels}Filtered channel features}

Before entering the experimental section, let us describe our general
architecture. Methods such as \texttt{ChnFtrs} \cite{Dollar2009Bmvc},
\texttt{SquaresChnFtrs} \cite{Benenson2013Cvpr,Benenson2014Eccvw}
and \texttt{ACF} \cite{Dollar2014Pami} all use the basic architecture
depicted in figure \ref{fig:Filtered-integral-channels-ilustration}
top part (best viewed in colours). The input image is transformed
into a set of feature channels (also called feature maps), the feature
vector is constructed by sum-pooling over a (large) set of rectangular
regions. This feature vector is fed into a decision forest learned
via Adaboost. The split nodes in the trees are a simple comparison
between a feature value and a learned threshold. Commonly only a subset
of the feature vector is used by the learned decision forest. Adaboost
serves both for feature selection and for learning the thresholds
in the split nodes.

A key observation, illustrated in figure \ref{fig:Filtered-integral-channels-ilustration}
(bottom), is that such sum-pooling can be re-written as convolution
with a filter bank (one filter per rectangular shape) followed by
reading a single value of the convolution's response map. This ``filter
+ pick'' view generalizes the integral channel features \cite{Dollar2009Bmvc}
detectors by allowing to use any filter bank (instead of only rectangular
shapes). We name this generalization ``filtered channel features
detectors''.

In our framework, \texttt{ACF} \cite{Dollar2014Pami} has a single
filter in its bank, corresponding to a uniform $4\negmedspace\times\negmedspace4\ \mbox{pixels}$
pooling region. \texttt{ChnFtrs} \cite{Dollar2009Bmvc} was a very
large (tens of thousands) filter bank comprised of random rectangular
shapes. \texttt{SquaresChnFtrs} \cite{Benenson2013Cvpr,Benenson2014Eccvw},
on the other hand, was only $16$ filters, each with a square-shaped
uniform pooling region of different sizes. See figure \ref{fig:ACF-SquaresChnFtrs-filters}
for an illustration of the \texttt{SquaresChnFtrs} filters, the upper-left
filter corresponds to \texttt{ACF}'s one.

The \texttt{InformedHaar} \cite{Zhang2014CvprInformedHaar} method
can also be seen as a filtered channel features detector, where the
filter bank (and read locations) are based on a human shape template
(thus the ``informed'' naming). \texttt{LDCF} \cite{Nam2014Nips}
is also a particular instance of this framework, where the filter
bank consists of PCA bases of patches from the training dataset.
In sections \ref{sec:How-many-filters} and \ref{sec:Additional-training-data}
we provide experiments revisiting some of the design decisions of
these methods.

Note that all the methods mentioned above (and in the majority of
experiments below) use only HOG+LUV feature channels%
\footnote{We use ``raw'' HOG, without any clamping, cell normalization, block
normalization, or dimensionality reduction.%
} (10 channels total). Using linear filters and decision trees on top
of these does not allow to reconstruct the decision functions obtained
when using LBP or covariance features (used by \texttt{SpatialPooling}
and \texttt{Regionlets}). We thus consider the approach considered
here orthogonal to adding such types of features.

\subsection{\label{sub:Evaluation-protocol}Evaluation protocol}

For our experiments we use the Caltech \cite{Dollar2011Pami,Benenson2014Eccvw}
and KITTI datasets \cite{Geiger2012CVPR}. The popular INRIA dataset
is considered too small and too close to saturation to provide interesting
results. All Caltech results are evaluated using the provided toolbox,
and summarised by log-average miss-rate (MR, lower is better) in the
$\left[10^{-2},\,10^{0}\right]\ \mbox{FPPI}$ range for the ``reasonable''
setup. KITTI results are evaluated via the online evaluation portal,
and summarised as average precision (AP, higher is better) for the
``moderate'' setup.

\paragraph{Caltech10x}

The raw Caltech dataset consists of videos (acquired at $30\,\mbox{Hz}$)
with every frame annotated. The standard training and evaluation considers
one out of each $30$ frames ($1\,631$ pedestrians over $4\,250$
frames in training, $1\,014$ pedestrians over $4\,024$ frames in
testing).\\
In our experiments of section \ref{sec:Additional-training-data}
we will also consider a $10\times$ increased training set where every
$3$rd frame is used (linear growth in pedestrians and images). We
name this extended training set ``Caltech10x''. \texttt{LDCF} \cite{Nam2014Nips}
uses a similar extended set for training its model (every $4$th frame).

\paragraph{Flow}

Methods using optical flow do not only use additional neighbour frames
during training ($1\leftrightarrow4$ depending on the method), but
they also do so at test time. Because they have access to additional
information at test time, we consider them as a separate group in
our results section.

\paragraph{Validation set}

In order to explore the design space of our pedestrian detector we
setup a Caltech validation set by splitting the six training videos
into five for training and one for testing (one of the splits suggested
in \cite{Dollar2011Pami}). Most of our experiments use this validation
setup. We also report (a posteriori) our key results on the standard
test set for comparison to the state of the art.\\
For the KITTI experiments we also validate some design choices (such
as search range and number of scales) before submission on the evaluation
server. There we use a $\nicefrac{2}{3}+\nicefrac{1}{3}$ validation
setup.

\subsection{\label{sub:Baselines}Baselines}

\paragraph{\texttt{ACF}}

Our experiments are based on the open source release of \texttt{ACF}
\cite{Dollar2014Pami}. Our first baseline is vanilla \texttt{ACF}
re-trained on the standard Caltech set (\emph{not} Caltech10x). On
the Caltech test set it obtains $32.6\%\ \mbox{MR}$ ($50.2\%\ \mbox{MR}$
on validation set). Note that this baseline already improves over
more than $50$ previously published methods \cite{Benenson2014Eccvw}
on this dataset. There is also a large gap between \texttt{ACF-Ours}
($32.6\%\ \mbox{MR}$) and the original number from \texttt{ACF-Caltech}
($44.2\%\ \mbox{MR}$ \cite{Dollar2014Pami}). The improvement is
mainly due to the change towards a larger model size (from $30\negmedspace\times\negmedspace60\ \mbox{pixels}$
to $60\negmedspace\times\negmedspace120$). All parameter details
are described in section \ref{sub:Training-parameters}, and kept
identical across experiments unless explicitly stated.

\paragraph{\texttt{InformedHaar}}

Our second baseline is a re-implementation of \texttt{InformedHaar}
\cite{Zhang2014CvprInformedHaar}. Here again we observe an important
gain from using a larger model size (same change as for \texttt{ACF}).
While the original \texttt{InformedHaar} paper reports $34.6\%\ \mbox{MR}$,
\texttt{Informed\-Haar-Ours} reaches $27.0\%\ \mbox{MR}$ on the
Caltech test set ($39.3\%\ \mbox{MR}$ on validation set). 

For both our baselines we use exactly the same training set as the
original papers. Note that the \texttt{Informed\-Haar-Ours} baseline
($27.0\%\ \mbox{MR}$) is right away the best known result for a method
trained on the standard Caltech training set. In section \ref{sec:Filter-banks-families}
we will discuss our re-implementation of \texttt{LDCF} \cite{Nam2014Nips}.

\subsection{\label{sub:Training-parameters}Model parameters}

Unless otherwise specified we train all our models using the following
parameters. Feature channels are HOG+LUV only. The final classifier
includes $4096$ level-2 decision trees (L2, 3 stumps per tree), trained
via vanilla discrete Adaboost. Each tree is built by doing exhaustive
greedy search for each node (no randomization). The model has size
$60\negmedspace\times\negmedspace120\ \mbox{pixels}$, and is built
via four rounds of hard negative mining (starting from a model with
$32$ trees, and then $512$, $1024$, $2048$, $4096$ trees). Each
round adds $10\,000$ additional negatives to the training set. The
sliding window stride is $6\ \mbox{pixels}$ (both during hard negative
mining and at test time).

Compared to the default \texttt{ACF} parameters, we use a bigger model,
more trees, more negative samples, and more boosting rounds. But we
do use the same code-base and the same training set. 

Starting from section \ref{sec:Additional-training-data} we will
also consider results with the Caltech10x data, there we use level-4
decision trees (L4), and Realboost \cite{Friedman2000} instead of
discrete Adaboost. All other parameters are left unchanged.

\section{\label{sec:Filter-banks-families}Filter bank families}

\begin{figure}
\begin{centering}
\begin{tabular}{cc}
\subfloat[\texttt{\label{fig:ACF-SquaresChnFtrs-filters}SquaresChntrs }filters]{\centering{}\includegraphics[width=0.2\textwidth]{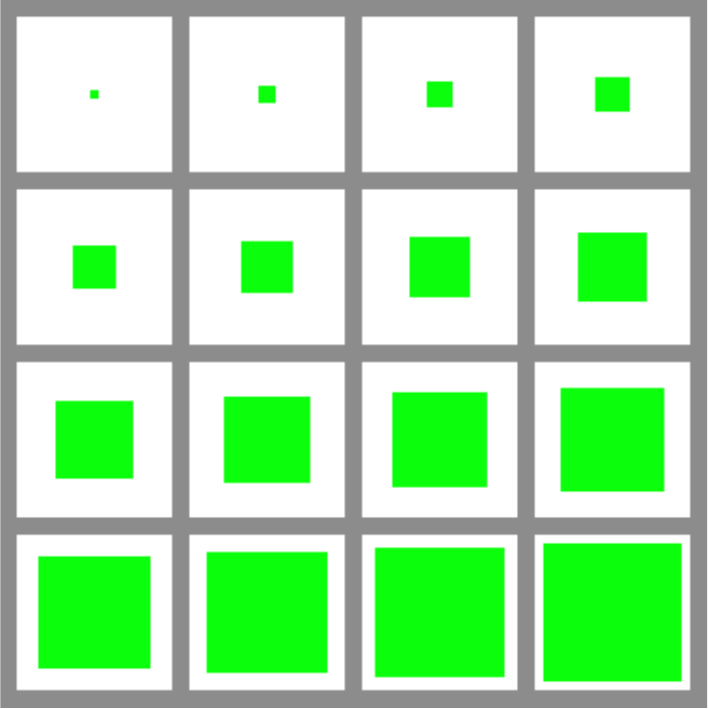}} & \subfloat[\texttt{\label{fig:CheckerBoard-filters}Checkerboards }filters]{\centering{}\includegraphics[width=0.2\textwidth]{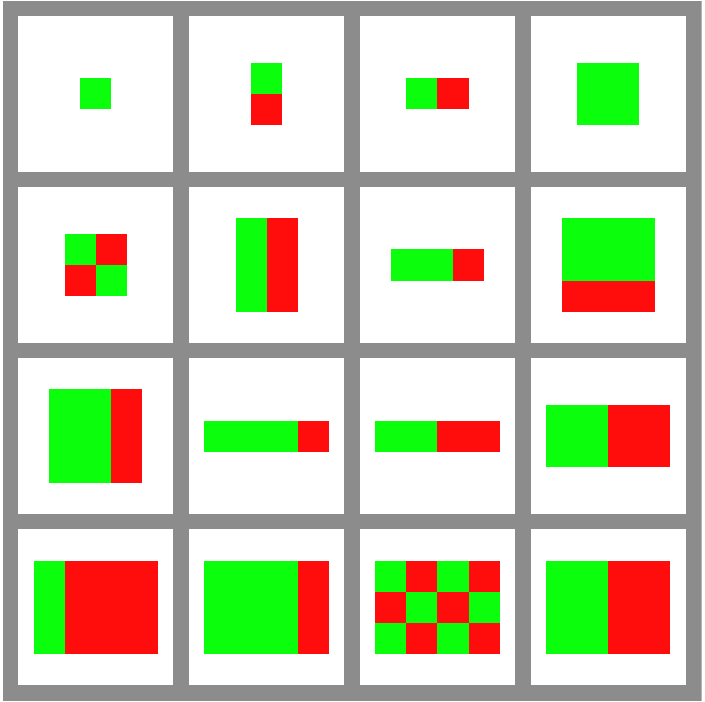}}\tabularnewline
\subfloat[\texttt{\label{fig:RandomFilters}RandomFilters}\quad{}]{\centering{}\includegraphics[width=0.2\textwidth]{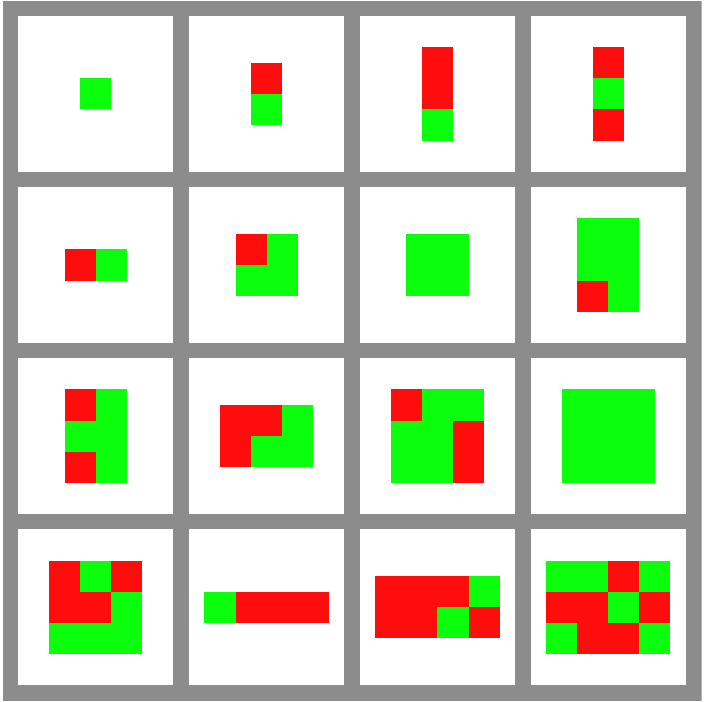}} & \subfloat[\texttt{\label{fig:InformedFilters}InformedFilters}]{\centering{}\includegraphics[width=0.2\textwidth]{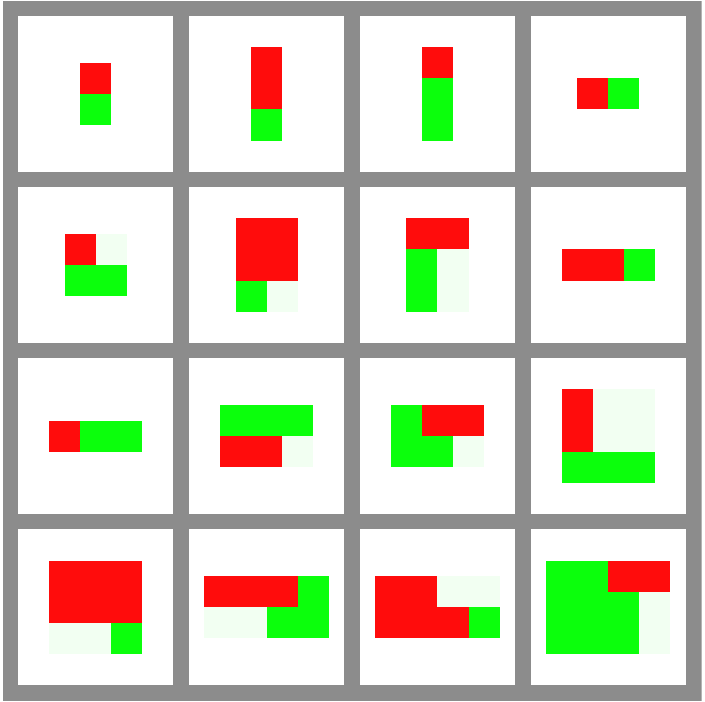}}\tabularnewline
\subfloat[\texttt{\label{fig:LDCF-filters}LDCF8} filters]{\centering{}\includegraphics[width=0.2\textwidth]{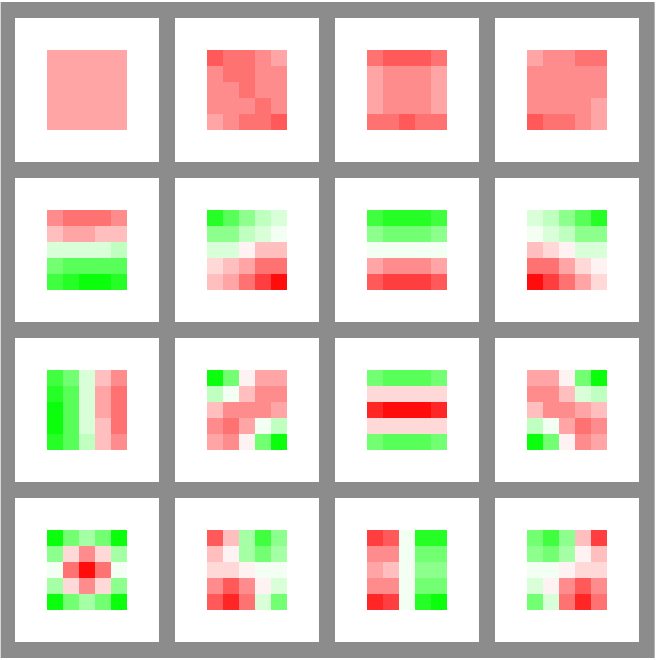}} & \subfloat[\texttt{\label{fig:PcaForeground-filters}PcaForeground} filters]{\centering{}\includegraphics[width=0.2\textwidth]{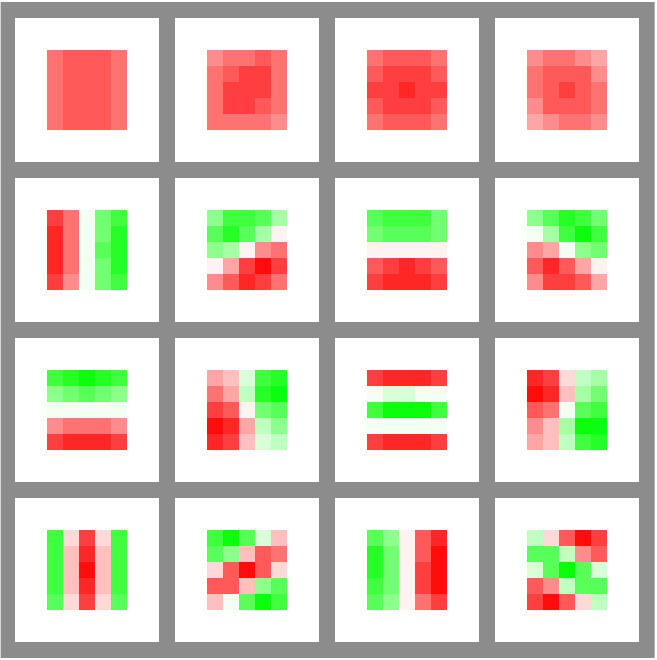}}\tabularnewline
\end{tabular}\vspace{-0.5em}

\par\end{centering}

\protect\caption{\label{fig:Filter-banks-illustration}Illustration of the different
filter banks considered. Except for \texttt{SquaresChntrs} filters,
only a random subset of the full filter bank is shown. \{${\color{green}{\color{red}\blacksquare}}$
Red,\foreignlanguage{english}{\textcolor{green}{{} ${\normalcolor \square}$}}
White,\foreignlanguage{english}{\textcolor{green}{{} ${\color{green}\blacksquare}$}}
Green\} indicate $\{-1,\,0,\,+1\}$.}
\vspace{-1em}
\end{figure}
Given the general architecture and the baselines described in section
\ref{sec:Filtered-integral-channels}, we now proceed to explore different
types of filter banks. Some of them are designed using prior knowledge
and they do not change when applied across datasets, others exploit
data-driven techniques for learning their filters. Sections \ref{sec:How-many-filters}
and \ref{sec:Additional-training-data} will compare their detection
quality.

\paragraph{\texttt{InformedFilters}}

Starting from the \texttt{Informed\-Haar} \cite{Zhang2014CvprInformedHaar}
baseline we use the same ``informed'' filters but let free the positions
where they are applied (instead of fixed in \texttt{Informed\-Haar});
these are selected during the boosting learning. Our initial experiments
show that removing the position constraint has a small (positive)
effect. Additionally we observe that the original \texttt{InformedHaar}
filters do not include simple square pooling regions (à la \texttt{SquaresChnFtrs}),
we thus add these too. We end up with $212$ filters in total, to
be applied over each of the $10$ feature channels. This is equivalent
to training decision trees over $2120$ (non filtered) channel features.\\
As illustrated in figure \ref{fig:InformedFilters} the \texttt{InformedFilters}
have different sizes, from $1\negmedspace\times\negmedspace1$ to
$4\negmedspace\times\negmedspace3$ cells ($1\mbox{ cell}=6\negmedspace\times\negmedspace6\ \mbox{pixels}$),
and each cell takes a value in $\left\{ -1,\,0,\,+1\right\} $. These
filters are applied with a step size of $6\ \mbox{pixels}$. For a
model of $60\negmedspace\times\negmedspace120\ \mbox{pixels}$ this
results in $200$ features per channel, $2\,120\cdot200=424\,000$
features in total%
\footnote{``Feature channel'' refers to the output of the first transformation
in figure \ref{fig:Filtered-integral-channels-ilustration} bottom.
``Filters'' are the convolutional operators applied to the feature
channels. And ``features'' are entries in the response maps of all
filters applied over all channels. A subset of these features are
the input to the learned decision forest.%
}. In practice considering border effects (large filters are not applied
on the border of the model to avoid reading outside it) we end up
with $\sim\negthickspace300\,000$ features. When training $4\,096$
level-2 decision trees, at most $4\,096\cdot3=12\,288$ features will
be used, that is $\sim\negthickspace3\%$ of the total. In this scenario
(and all others considered in this paper) Adaboost has a strong role
of feature selection.

\paragraph{\texttt{Checkerboards}}

As seen in section \ref{sub:Baselines} \texttt{InformedHaar} is a
strong baseline. It is however unclear how much the ``informed''
design of the filters is effective compared to other possible choices.
\texttt{Checker\-boards} is a naïve set of filters that covers the
same sizes (in number of cells) as \texttt{Informed\-Haar/Informed\-Filters}
and for each size defines (see figure \ref{fig:CheckerBoard-filters}):
a uniform square, all horizontal and vertical gradient detectors ($\pm1$
values), and all possible checkerboard patterns. These configurations
are comparable to \texttt{InformedFilters} but do not use the human
shape as prior.\\
The total number of filters is a direct function of the maximum size
selected. For up to $4\negmedspace\times\negmedspace4$ cells we end
up with $61$ filters, up to $4\negmedspace\times\negmedspace3$ cells
$39$ filters, up to $3\negmedspace\times\negmedspace3$ cells $25$
filters, and up to $2\negmedspace\times\negmedspace2$ cells $7$
filters.

\paragraph{\texttt{RandomFilters}}

Our next step towards removing a hand-crafted design is simply using
random filters (see figure \ref{fig:RandomFilters}). Given a desired
number of filters and a maximum filter size (in cells), we sample
the filter size with uniform distribution, and set its cell values
to $\pm1$ with uniform probability. We also experimented with values
$\{-1,\,0,\,+1\}$ and observed a (small) quality decrease compared
to the binary option).

The design of the filters considered above completely ignores the
available training data. In the following, we consider additional
filters learned from data. 
\begin{figure}
\begin{centering}
\includegraphics[width=1\columnwidth]{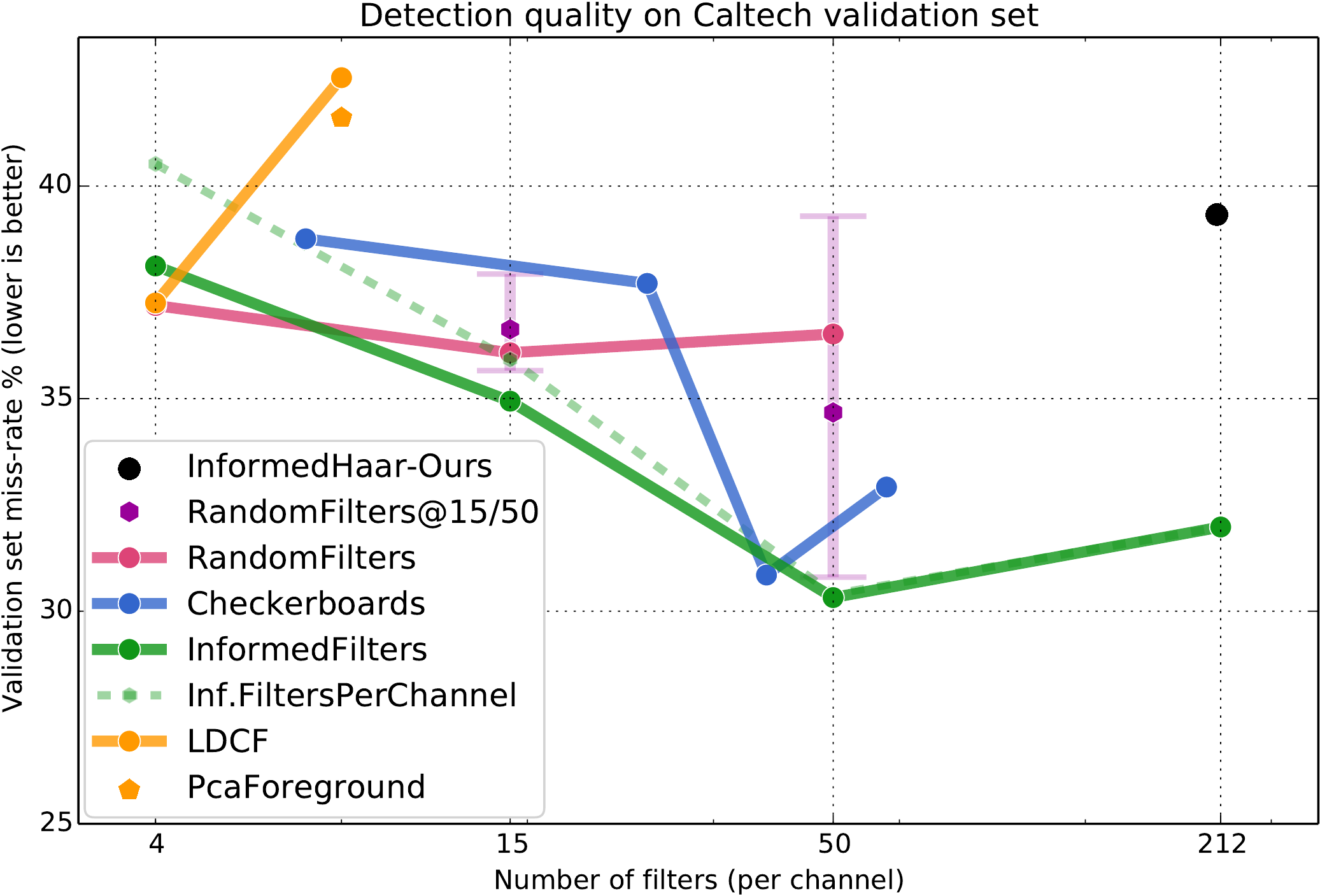}\vspace{-0.5em}

\par\end{centering}

\protect\caption{\label{fig:filters-quality-small-data}Detection quality (log-average
miss-rate MR, lower is better) versus number of filters used. All
models trained and tested on the Caltech validation set (see \S\ref{sec:How-many-filters}).}
\vspace{-1em}
\end{figure}

\paragraph{\texttt{LDCF} \cite{Nam2014Nips}}

The work on PCANet \cite{Chan2014ArxivPcanet} showed that applying
arbitrary non-linearities on top of PCA projections of image patches
can be surprisingly effective for image classification. Following
this intuition \texttt{LDCF} \cite{Nam2014Nips} uses learned PCA
eigenvectors as filters (see figure \ref{fig:LDCF-filters}).\\
We present a re-implementation of \cite{Nam2014Nips} based on \texttt{ACF}'s
\cite{Dollar2014Pami} source code. We try to follow the original
description as closely as possible. We use the same top $4$ filters
of $10\negmedspace\times\negmedspace10\ \mbox{pixels}$, selected
per feature channel based on their eigenvalues ($40$ filters total).
We do change some parameters to be consistent amongst all experiments,
see sections \ref{sub:Training-parameters} and \ref{sec:Additional-training-data}.
The main changes are the training set (we use Caltech10x, sampled
every 3 frames, instead of every 4 frames in \cite{Nam2014Nips}),
and the model size ($60\negmedspace\times\negmedspace120$ pixels
instead of $32\negmedspace\times\negmedspace64$). As will be shown
in section \ref{sec:Test-set-results}, our implementation (\texttt{LDCF-Ours})
clearly improves over the previously published numbers \cite{Nam2014Nips},
showing the potential of the method.\\
For comparison with \texttt{PcaForeground} we also consider training
\texttt{LDCF8} where the top $8$ filters are selected per channel
($80$ filters total).

\paragraph{\texttt{PcaForeground}}

In \texttt{LDCF} the filters are learned using all of the training
data available. In practice this means that the learned filters will
be dominated by background information, and will have minimal information
about the pedestrians. Put differently, learning filters from all
the data assumes that the decision boundary is defined by a single
distribution (like in Linear Discriminant Analysis \cite{Murphy2012MachineLearningBook}),
while we might want to define it based on the relation between the
background distribution and the foreground distribution (like Fisher's
Discriminant Analysis \cite{Murphy2012MachineLearningBook}). In \texttt{PcaForeground}
we train $8$ filters per feature channel, $4$ learned from background
image patches, and $4$ learned from patches extracted over pedestrians
(see figure \ref{fig:PcaForeground-filters}). Compared to \texttt{LDCF8}
the obtained filters are similar but not identical, all other parameters
are kept identical.\\
Other than via \texttt{PcaForeground}/\texttt{LDCF8, }it is not clear
how to further increase the number of filters used in \texttt{LDCF}.
Past $8$ filters per channel, the eigenvalues decrease to negligible
values and the eigenvectors become essentially random (similar to
\texttt{RandomFilters}).

To keep the filtered channel features setup close to \texttt{Infor\-med\-Haar},
the filters are applied with a step of $6\ \mbox{pixels}.$ However,
to stay close to the original \texttt{LDCF}, the \texttt{LDCF}/\texttt{Pca\-Fore\-ground}
filters are evaluated every $2\ \mbox{pixels}$. Although (for example)
\texttt{LDCF8} uses only $\sim\negthickspace10\%$ of  the number
of filters per channel compared to \texttt{Che\-cker\-boards4x4},
due to the step size increase, the obtained feature vector size is
$\sim\negthickspace40\%$.

\section{\label{sec:How-many-filters}How many filters?}

Given a fixed set of channel features, a larger filter bank provides
a richer view over the data compared to a smaller one. With enough
training data one would expect larger filter banks to perform best.
We want thus to analyze the trade-off between number of filters and
detection quality, as well as which filter bank family performs best.

Figure \ref{fig:filters-quality-small-data} presents the results
of our initial experiments on the Caltech validation set. It shows
detection quality versus number of filters per channel. This figure
densely summarizes $\sim\negthickspace30$ trained models.

\paragraph{\texttt{InformedFilters}}

The first aspect to notice is that there is a meaningful gap between
\texttt{Informed\-Haar-Ours }and \texttt{Informed\-Filters} despite
having a similar number of filters ($209$ versus $212$). This validates
the importance of letting Ada\-boost choose the pooling locations
instead of hand-crafting them. Keep in mind that \texttt{In\-for\-med\-Haar-Ours}
is a top performing baseline (see \S\ref{sub:Baselines}).\\
Secondly, we observe that (for the fixed training data available)
$\sim\negthickspace50$ filters is better than $\sim\negthickspace200$.
Below $50$ filters the performance degrades for all methods (as expected).\\
To change the number of filters in \texttt{Informed\-Filters} we
train a full model ($212$ filters), pick the $N$ most frequently
used filters (selected from node splitting in the decision forest),
and use these to train the desired reduced model. We can select the
most frequent filters across channels or per channel (marked as \texttt{Inf.FiltersPerChannel}).
We observe that per channel selection is slightly worse than across
channels, thus we stick to the latter.\\
Using the most frequently used filters for selection is clearly a
crude strategy since frequent usage does not guarantee discriminative
power, and it ignores relation amongst filters. We find this strategy
good enough to convey the main points of this work.
\begin{table}
\centering{}%
\begin{tabular}{cc|cccc}
Training & Method & $\mbox{L}2$ & $\mbox{L}3$  & $\mbox{L}4$  & $\mbox{L}5$\tabularnewline
\hline 
\hline 
Caltech & \multirow{2}{*}{\texttt{ACF}} & $50.2$ & \emph{$\mathit{42.1}$} & $48.8$ & $48.7$\tabularnewline
Caltech10x &  & $52.6$ & $49.9$ & \emph{$44.9$} & $\mathit{41.3}$\tabularnewline
\hline 
Caltech & \texttt{Checker-} & $32.9$ & $30.4$ & \emph{$\mathit{28.0}$} & $31.5$\tabularnewline
Caltech10x & \texttt{boards} & $37.0$ & $31.6$ & $\mathit{24.7}$ & \emph{$\mathit{24.7}$}\tabularnewline
\end{tabular}\vspace{-0.5em}
\protect\caption{\label{tab:Training-volume}Effect of the training volume and decision
tree depth ($\mbox{L}n$) over the detection quality (average miss-rate
on validation set, lower is better), for \texttt{ACF-\-Ours }and
\texttt{Checker\-boards} variant with ($61$) filters of $4\negmedspace\times\negmedspace4\ \mbox{cells}$.
We observe a similar trend for other filter banks. }
\vspace{-1em}
\end{table}

\paragraph{\texttt{Checkerboards}}

also reaches best results in the $\sim\negthickspace50$ filters region.
Here the number of filters is varied by changing the maximum filter
size (in number of cells). Regarding the lowest miss-rate there is
no large gap between the ``informed'' filters and this naïve baseline.

\paragraph{\texttt{RandomFilters}}

The hexagonal dots and their deviation bars indicate the mean, maximum
and minimum miss-rate obtained out of five random runs. When using
a larger number of filters ($50$) we observe a lower (better) mean
but a larger variance compared to when using fewer filters ($15$).
Here again the gap between the best random run and the best result
of other methods is not large.\\
Given a set of five models, we select the $N$ most frequently used
filters and train new reduced models; these are shown in the \texttt{Random\-Filters
}line. Overall the random filters are surprisingly close to the other
filter families. This indicates that expanding the feature channels
via filtering is the key step for improving detection quality, while
selecting the ``perfect'' filters is a secondary concern.

\paragraph{\texttt{LDCF}/\texttt{PcaForeground}}

In contrast to the other filter bank families, \texttt{LDCF} under-performs
when increasing the number of filters (from $4$ to $8$) while using
the standard Caltech training set (consistent with the observations
in \cite{Nam2014Nips}). \texttt{PcaForeground} improves marginally
over \texttt{LDCF8}.

\paragraph{Takeaways}

From figure \ref{fig:filters-quality-small-data} we observe two overall
trends. First, the more filters the merrier, with $\sim\negthickspace50$
filters as sweet spot for Caltech training data. Second, there is
no flagrant difference between the different filter types.

\section{\label{sec:Additional-training-data}Additional training data}

One caveat of the previous experiments is that as we increase the
number of filters used, so does the number of features Adaboost must
pick from. Since we increased the model capacity (compared to \texttt{ACF}
which uses a single filter), we consider using the Caltech10x dataset
(\S\ref{sub:Evaluation-protocol}) to verify that our models are
not starving for data. Similar to the experiments in \cite{Nam2014Nips},
we also reconsider the decision tree depth, since additional training
data enables bigger models. 

Results for two representative methods are collected in table \ref{tab:Training-volume}.
First we observe that already with the original training data, deeper
trees do provide significant improvement over level-2 (which was selected
when tuning over INRIA data \cite{Dollar2009Bmvc,Benenson2013Cvpr}).
Second, we notice that increasing the training data volume does provide
the expected improvement only when the decision trees are deep enough.
For our following experiments we choose to use level-4 decision trees
($\mbox{L}4$) as a good balance between increased detection quality
and reasonable training times.

\paragraph{Realboost}

Although previous papers on \texttt{ChnFtrs} detectors reported that
different boosting variants all obtain equal results on this task
\cite{Dollar2009Bmvc,Benenson2013Cvpr}, the recent \cite{Nam2014Nips}
indicated that Realboost has an edge over discrete Adaboost when additional
training data is used. We observe the same behaviour in our Caltech10x
setup. 
\begin{table}
\begin{centering}
\begin{tabular}{lcc}
Aspect & MR  & $\Delta\mbox{MR}$\tabularnewline
\hline 
\hline 
\texttt{ACF-Ours} & $50.8$ & -\tabularnewline
\hline 
+ filters & $32.9$ & $+17.9$\tabularnewline
+ L4 & $28.0$ & $+4.9$\tabularnewline
+ Caltech10x  & \selectlanguage{english}%
$24.7$\selectlanguage{british}%
 & $+3.3$\tabularnewline
+ Realboost & $24.4$ & $+0.3$\tabularnewline
\hline 
\hline 
\texttt{\small{}Checker\-boards4x4} & $24.4$ & $+26.4$\tabularnewline
\end{tabular}\vspace{-0.5em}

\par\end{centering}

\centering{}\protect\caption{\label{tab:Ingredients-strong-detector}Ingredients to build our strong
detectors (using \texttt{Checker\-boards4x4} in this example, 61
filters). Validation set log-average miss-rate (MR).}
\vspace{-1em}
\end{table}

As summarized in table \ref{tab:Ingredients-strong-detector} using
filtered channels, deeper trees, additional training data, and Realboost
does provide a significant detection quality boost. For the rest of
the paper our models trained on Caltech10x all use level-4 trees and
RealBoost, instead of level-2 and discrete Adaboost for the Caltech1x
models.

\paragraph{Timing}

When using Caltech data \texttt{ACF} takes about one hour for training
and one for testing. \texttt{Che\-cker\-boards\-4x4} takes about
$4$ and $2$ hours respectively. When using Caltech10x the training
times for these methods augment to 2 and 29 hours, respectively. The
training time does not increase proportionally with the training data
volume because the hard negative mining reads a variable amount of
images to attain the desired quota of negative samples. This amount
increases when a detector has less false positive mistakes.

\subsection{\label{sub:Validation-set-experiments}Validation set experiments}

\begin{table}
\begin{centering}
\begin{tabular}{c|c|ccc}
\multirow{2}{*}{Filters type} & {\footnotesize{}\#} & {\footnotesize{}Caltech} & {\footnotesize{}Caltech10x} & \multirow{2}{*}{$\Delta\mbox{MR}$}\tabularnewline
 & {\footnotesize{}filters} & MR & MR & \tabularnewline
\hline 
\hline 
\texttt{\small{}ACF-Ours} & $1$ & $50.2$ & $39.8$ & $10.4$\tabularnewline
\texttt{\small{}LDCF-Ours} & $4$ & $37.3$ & $34.1$ & $3.2$\tabularnewline
\texttt{\small{}LDCF8} & $8$ & $42.6$ & $30.7$ & $11.9$\tabularnewline
\texttt{\small{}PcaForeground} & $8$ & $41.6$ & $28.6$ & $13.0$\tabularnewline
\texttt{\small{}RandomFilters} & $50$ & $36.5$ & $28.2$ & $8.3$\tabularnewline
\texttt{\small{}InformedFilters} & $50$ & $30.3$ & $26.6$ & $3.7$\tabularnewline
\texttt{\small{}Checkerboards} & $39$ & $30.9$ & $25.9$ & $5.0$\tabularnewline
\texttt{\small{}Checkerboards} & $61$ & $32.9$ & $24.4$ & $8.5$\tabularnewline
\end{tabular}\vspace{-0.5em}

\par\end{centering}

\centering{}\protect\caption{\label{tab:1x-vs-10x-valtech-results}Effect of increasing the training
set for different methods, quality measured on Caltech validation
set (MR: log-average miss-rate).}
\vspace{-1em}
\end{table}

Based on the results in table \ref{tab:Ingredients-strong-detector}
we proceed to evaluate on Caltech10x the most promising configurations
(filter type and number) from section \ref{sec:How-many-filters}.
The results over the Caltech validation set are collected in table
\ref{tab:1x-vs-10x-valtech-results}. We observe a clear overall gain
from increasing the training data.

Interestingly with enough \texttt{RandomFilters} we can outperform
the strong performance of \texttt{LDCF-Ours}. We also notice that
the naïve \texttt{Checkerboards} outperforms the manual design of
\texttt{InformedFilters}.

\section{\label{sec:All-in-one-methods}Add-ons}

Before presenting the final test set results of our ``core'' method
(section \ref{sec:Test-set-results}), we also consider some possible
``add-ons'' based on the suggestions from \cite{Benenson2014Eccvw}.
For the sake of evaluating complementarity, comparison with existing
method, and reporting the best possible detection quality, we consider
extending our detector with context and optical flow information.

\paragraph{Context}

Context is modelled via the \texttt{2Ped} re-scoring method of \cite{Ouyang2013Cvpr}.
It is a post-processing step that merges our detection scores with
the results of a two person DPM \cite{Felzenszwalb2010Pami} trained
on the INRIA dataset (with extended annotations).\\
In \cite{Ouyang2013Cvpr} the authors reported an improvement of $\sim\negthickspace5\ \mbox{pp}$
(percent points) on the Caltech set, across different methods. In
\cite{Benenson2014Eccvw} an improvement of $2.8\ \mbox{pp}$ is reported
over their strong detector (\texttt{SquaresChnFtrs+DCT+SDt} $25.2\%\ \mbox{MR}$).
In our experiments however we obtain a gain inferior to $0.5\ \mbox{pp}$.\\
We have also investigated fusing the \texttt{2Ped} detection results
via a different, more principled, fusion method \cite{Xu2014Bmvc}.
We observe consistent results: as the strength of the starting point
increases, the gain from \texttt{2Ped} decreases. When reaching our
\texttt{Checkerboards} results, all gains have evaporated. We believe
that the \texttt{2Ped} approach is a promising one, but our experiments
indicate that the used DPM template is simply too weak in comparison
to our filtered channels.

\paragraph{Optical flow}

Optical flow is fed to our detector as an additional set of $2$ channels
(not filtered). We use the implementation from \texttt{SDt} \cite{Park2013Cvpr}
which uses differences of weakly stabilized video frames. On Caltech\texttt{,}
the authors of \cite{Park2013Cvpr} reported a $\sim\negthickspace7\ \mbox{pp}$
gain over \texttt{ACF} ($44.2\%\ \mbox{MR}$), while \cite{Benenson2014Eccvw}
reported a $\sim\negthickspace5\ \mbox{pp}$ percent points improvement
over their strong baseline \texttt{(SquaresChnFtrs+DCT+2Ped }$27.4\%\ \mbox{MR}$\texttt{)}.
When using \texttt{+SDt} our results are directly comparable to \texttt{Katamari}
\cite{Benenson2014Eccvw} and \texttt{SpatialPooling+} \cite{Paisitkriangkrai2014Eccv}
which both use optical flow too.\\
Using our stronger \texttt{Checkerboards} results \texttt{SDt }provides
a $1.4\ \mbox{pp}$ gain. Here again we observe an erosion as the
starting point improves (for confirmation, reproduced the \texttt{ACF+SDt}
results \cite{Park2013Cvpr}, $43.9\%\!\rightarrow\!33.9\%\ \mbox{MR}$).
We name our \texttt{Checkerboards+SDt }detector \texttt{All-in-one}.

Our filtered channel features results are strong enough to erode existing
context and flow features. Although these remain complementary cues,
more sophisticated ways of extracting this information will be required
to further progress in detection quality.

It should be noted that despite our best efforts we could not reproduce
the results from neither \texttt{2Ped} nor \texttt{SDt} on the KITTI
dataset (in spite of its apparent similarity to Caltech). Effective
methods for context and optical flow across datasets have yet to be
shown. Our main contribution remains on the core detector (only HOG+LUV
features over local sliding window pixels in a single frame).

\section{\label{sec:Test-set-results}Test set results}

\begin{figure}
\begin{centering}
\includegraphics[width=0.9\columnwidth]{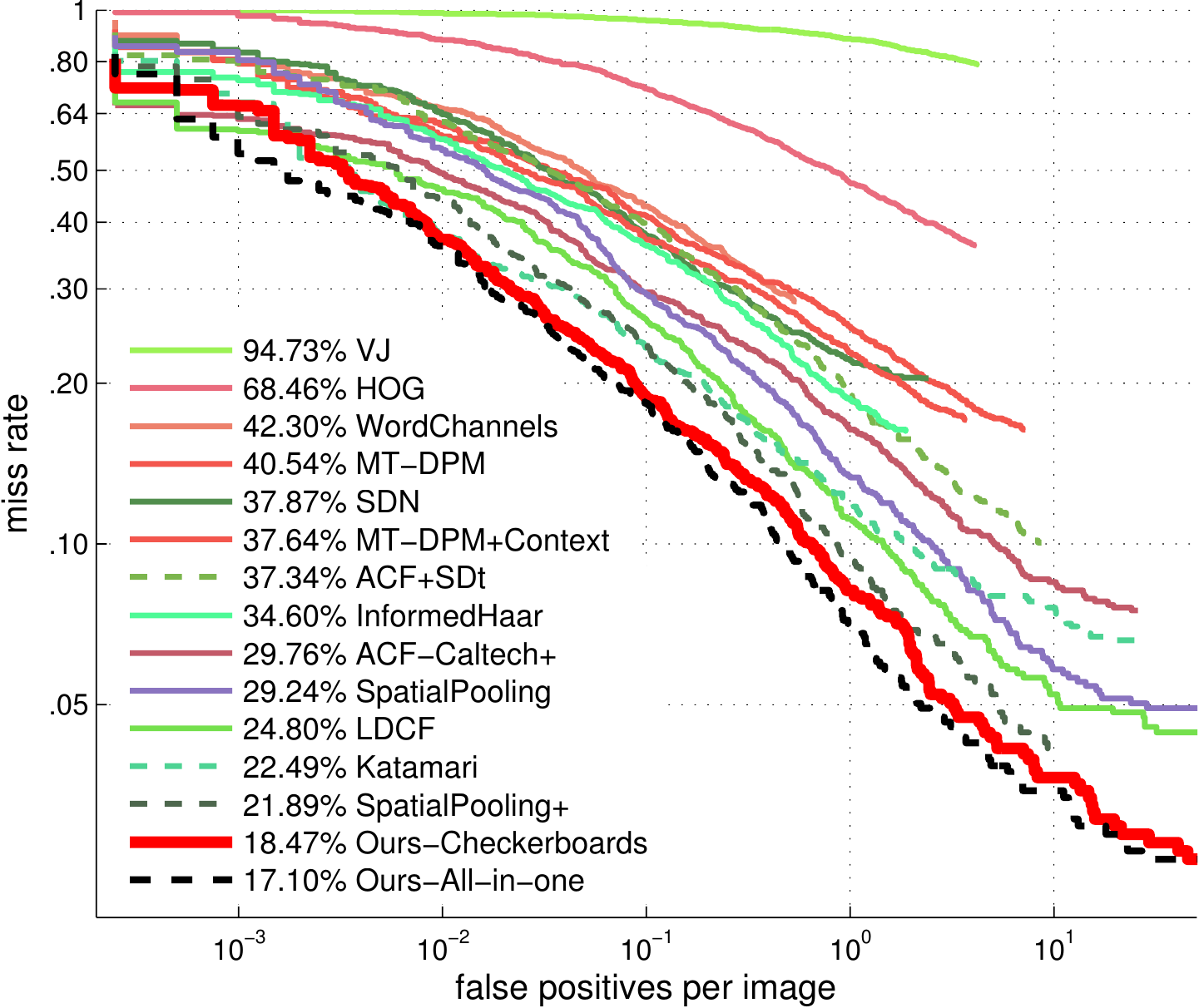}\vspace{-1em}

\par\end{centering}

\protect\caption{\label{fig:caltech-top-methods}Some of the top quality detection
methods for Caltech-USA.}
\vspace{-1em}
\end{figure}
Having done our exploration of the parameters space on the validation
set, we now evaluate the most promising methods on the Caltech and
KITTI test sets.

\paragraph{Caltech test set}

\begin{figure}
\begin{centering}
\includegraphics[width=1\columnwidth]{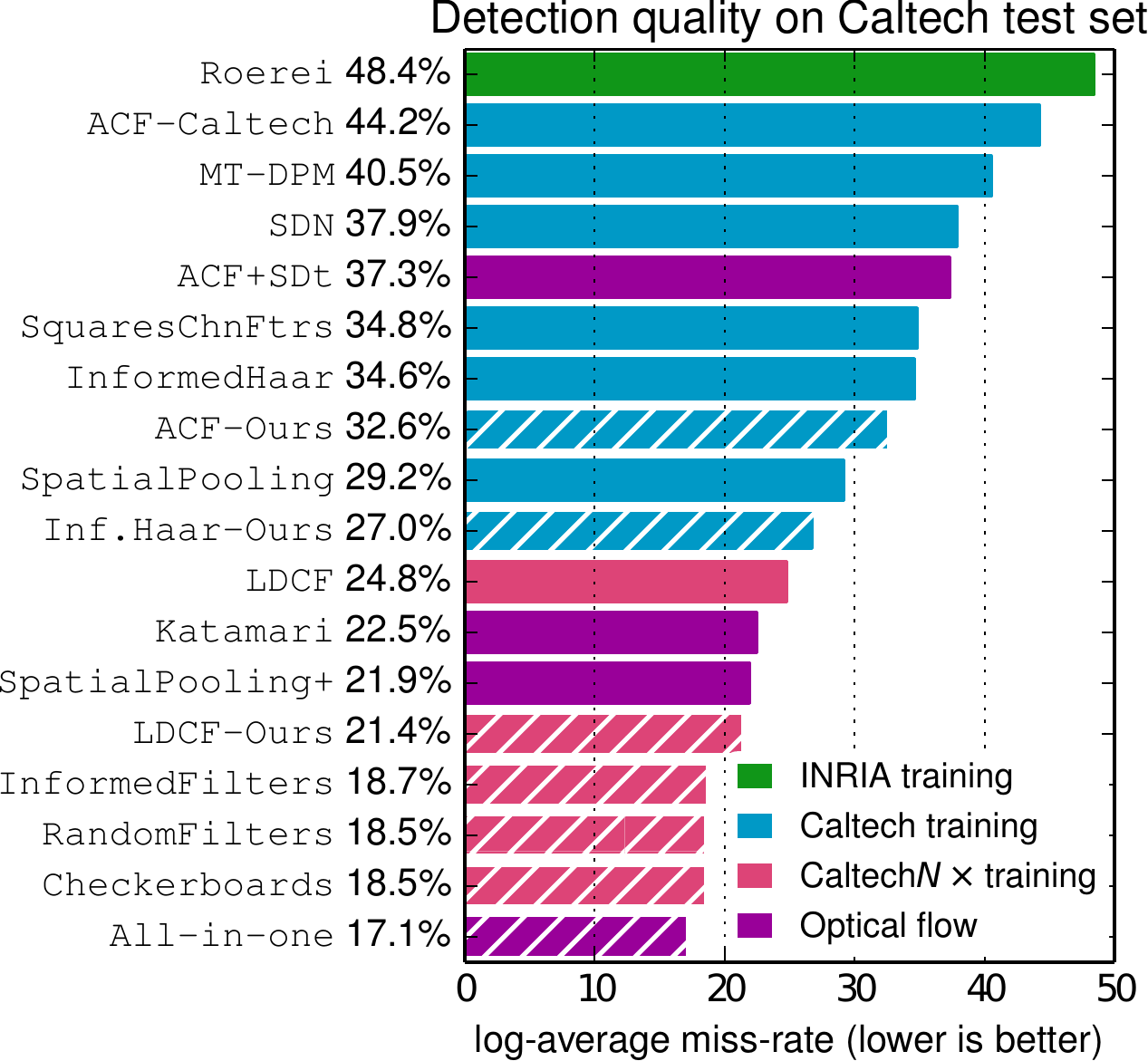}\vspace{-0.5em}

\par\end{centering}

\protect\caption{\label{fig:caltech-barplot}Some of the top quality detection methods
for Caltech test set (see text), and our results (highlighted with
white hatch). Methods using optical flow are trained on original Caltech
except our \texttt{All-\-in-\-one} which uses Caltech10x. $\mbox{Caltech}N\times$
indicates Caltech10x for all methods but the original \texttt{LDCF}
(see section \ref{sub:Evaluation-protocol}).}
\vspace{-1em}
\end{figure}

Figures \ref{fig:caltech-barplot} and \ref{fig:caltech-top-methods}
present our key results on the Caltech test set. For proper comparison,
only methods using the same training set should be compared (see \cite[figure 3]{Benenson2014Eccvw}
for a similar table comparing $50+$ previous methods). We include
for comparison the baselines mentioned in section \ref{sub:Baselines},
\texttt{Roerei \cite{Benenson2013Cvpr}} the best known method trained
without any Caltech images, \texttt{MT-DPM \cite{Yan2013Cvpr}} the
best known method based on DPM, and \texttt{SDN \cite{Luo2014Cvpr}}
the best known method using convolutional neural networks. We also
include the top performers \texttt{Katamari} \cite{Benenson2014Eccvw}
and \texttt{SpatialPooling+} \cite{Paisitkriangkrai2014Eccv}. We
mark as ``$\mbox{Caltech}N\times$'' both the Caltech10x training
set and the one used in \texttt{LDCF} \cite{Nam2014Nips} (see section
\ref{sec:Additional-training-data}).

\paragraph{KITTI test set}

Figure \ref{fig:kitti-results} presents the results on the KITTI
test set (``moderate'' setup), together with all other reported
methods using only monocular image content (no stereo or LIDAR data).
The KITTI evaluation server only recently has started receiving submissions
($14$ for this task, $11$ in the last year), and thus is less prone
to dataset over-fitting.\\
We train our model on the KITTI training set using almost identical
parameters as for Caltech. The only change is a subtle pre-processing
step in the HOG+LUV computation. On KITTI the input image is smoothed
(radius $1\ \mbox{pixel}$) before the feature channels are computed,
while on Caltech we do not. This subtle change provided a $\sim\negthickspace4\ \mbox{pp}$
(percent points) improvement on the KITTI validation set.

\subsection{Analysis}

With a $\sim\negthickspace10\ \mbox{pp}$ (percent points) gap between
\texttt{ACF}/\texttt{In\-for\-med\-Haar} and \texttt{ACF}/\texttt{In\-for\-med\-Haar\--Ours}
(see figure \ref{fig:caltech-barplot}), the results of our baselines
show the importance of proper validation of training parameters (large
enough model size and negative samples). \texttt{In\-for\-med\-Haar\--Ours}
is the best reported result when training with Caltech1x.

When considering methods trained on Caltech10x, we obtain a clear
gap with the previous best results (\texttt{LDCF} $24.8\%\ \mbox{MR}\rightarrow$\texttt{
Checker\-boards} $18.5\%\ \mbox{MR}$). Using our architecture and
the adequate number of filters one can obtain strong results using
only HOG+LUV features. The exact type of filters seems not critical,
in our experiments \texttt{Checker\-boards4x3} gets best performance
given the available training data. \texttt{RandomFilters} reaches
the same result, but requires training and merging multiple models.

Our results cut by half miss-rate of the best known \texttt{convnet}
for pedestrian detection (\texttt{SDN \cite{Luo2014Cvpr}}), which
in principle could learn similar low-level features and their filtering. 

When adding optical flow we further push the state of the art and
reach $17.1\%\ \mbox{MR}$, a comfortable $\sim\negthickspace5\ \mbox{pp}$
improvement over the previous best optical flow method (\texttt{Spa\-tial\-Pool\-ing}+).
This is the best reported result on this challenging dataset.

\begin{figure}
\begin{centering}
\includegraphics[width=0.9\columnwidth]{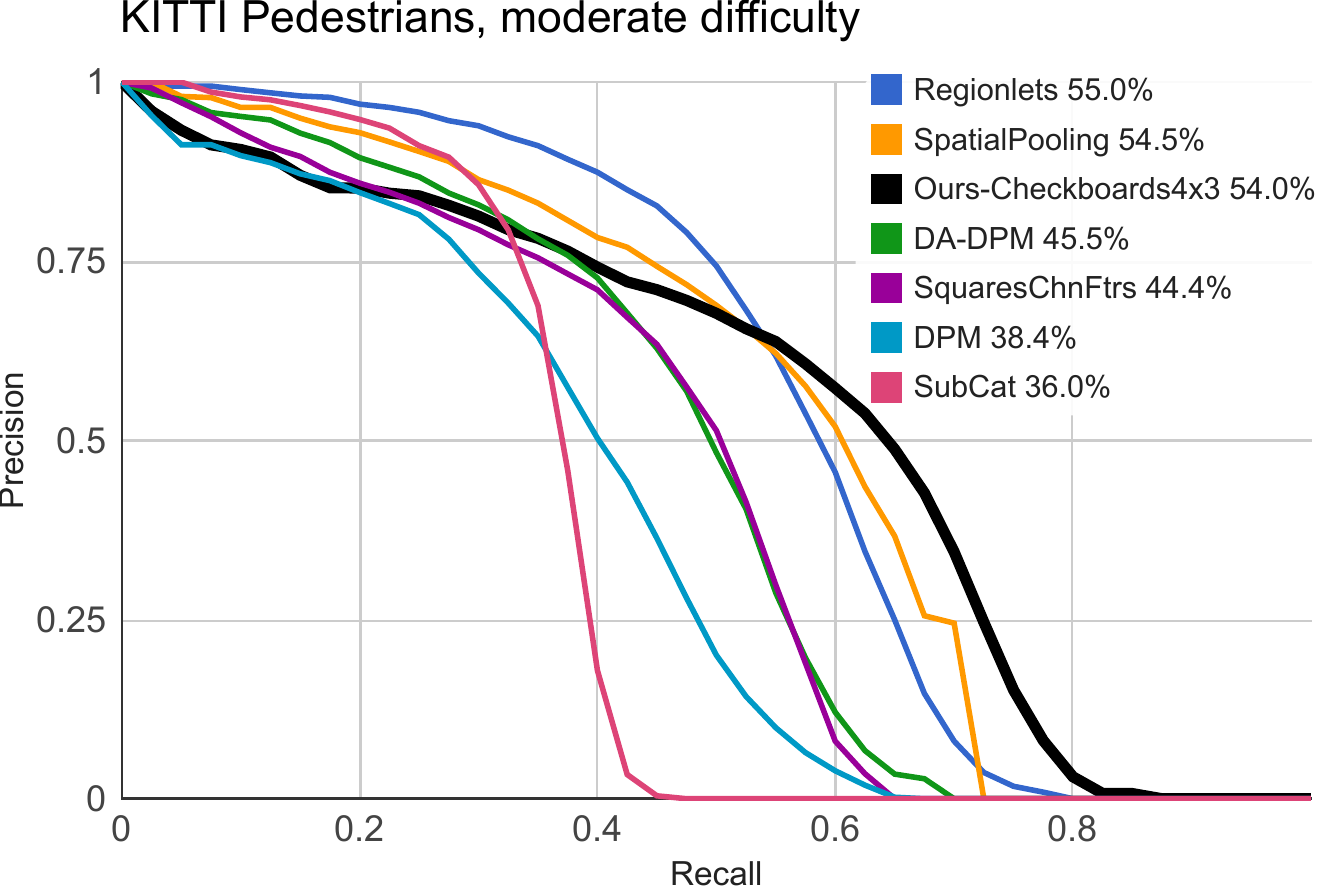}
\par\end{centering}

\begin{centering}
\vspace{-1em}

\par\end{centering}

\protect\caption{\label{fig:kitti-results}Pedestrian detection on the KITTI dataset
(using images only). }
\vspace{-1em}
\end{figure}

The results on the KITTI dataset confirm the strength of our approach,
reaching $54.0\%\ \mbox{AP}$, just $1\ \mbox{pp}$ below the best
known result on this dataset. Competing methods (\texttt{Regionlets
}\cite{Wang2013IccvRegionlets} and \texttt{SpatialPooling} \cite{Paisitkriangkrai2014Eccv})
both use HOG together with additional LBP and covariance features.
Adding these remains a possibility for our system. Note that our results
also improve over methods using LIDAR + Image, such as \texttt{Fusion-DPM}
\cite{Premebida2014Iros} ($46.7\%\ \mbox{AP}$, not included in figure
\ref{fig:kitti-results} for clarity).

\section{\label{sec:Conclusion}Conclusion}

Through this paper we have shown that the seemingly disconnected
methods \texttt{ACF}, \texttt{(Squares)ChnFtrs}, \texttt{InformedHaar},
and \texttt{LDCF} can be all put under the filtered channel features
detectors umbrella. We have systematically explored different filter
banks for such architecture and shown that they provide means for
important improvements for pedestrian detection. Our results indicate
that HOG+LUV features have not yet saturated, and that competitive
results (over Caltech and KITTI datasets) can be obtained using only
them. When optical flow information is added we set the new state
of art for the Caltech dataset, reaching $17.1\%\ \mbox{MR}$ ($93\%$
recall at $1$ false positive per image).

In future work we plan to explore how the insights of this work can
be exploited into a more general detection architecture such as convolutional
neural networks.

\paragraph{Acknowledgements}

We thank Jan Hosang for the help provided setting up some of the experiments.
We also thank Seong Joon Oh and Sabrina Hoppe for their useful comments.

\bibliographystyle{ieee}
\bibliography{2015_cvpr_filtered_channels_for_pedestrian_detection}

\appendix

\section{Learned model}

\begin{figure*}
\begin{centering}
\subfloat[{\label{fig:supp-Filters-from-Roerei}Filters from \texttt{Roerei }(scale
$1$) model. Copied from \cite[supplementary material]{Benenson2013Cvpr}}]{\begin{centering}
\includegraphics[height=8em]{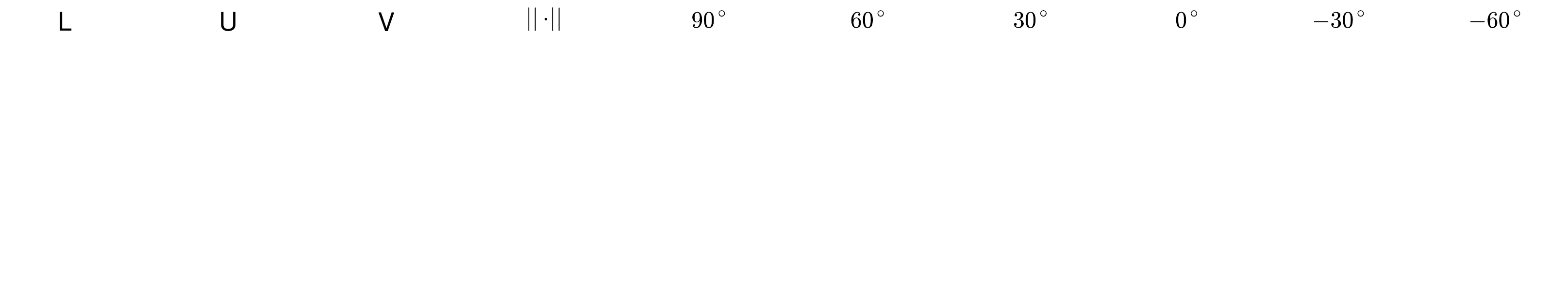}\qquad{}\includegraphics[height=8em]{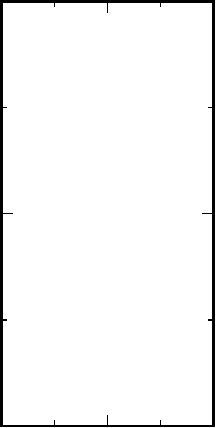}
\par\end{centering}

}
\par\end{centering}

\begin{centering}
\subfloat[\label{fig:supp-Checkerboards4x3-model}Final \texttt{Checker\-boards4x3
}model]{\begin{centering}
\includegraphics[height=8em]{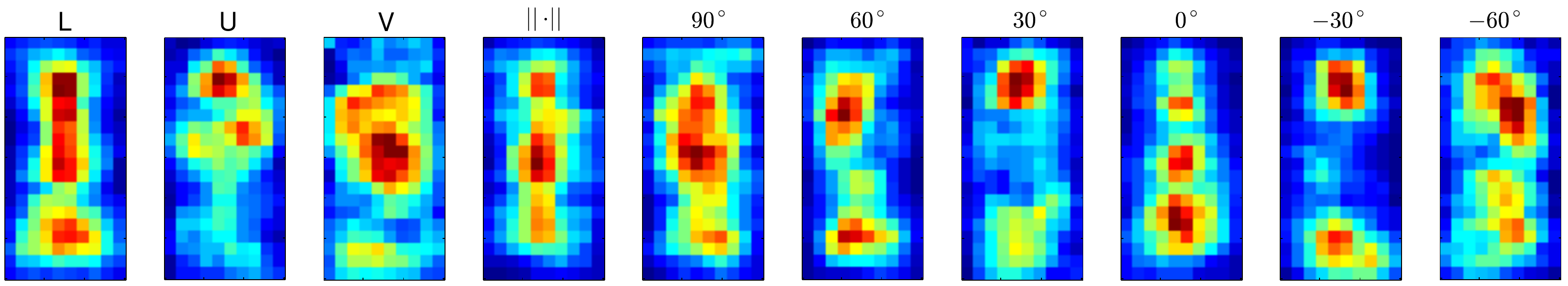}\qquad{}\includegraphics[height=8em]{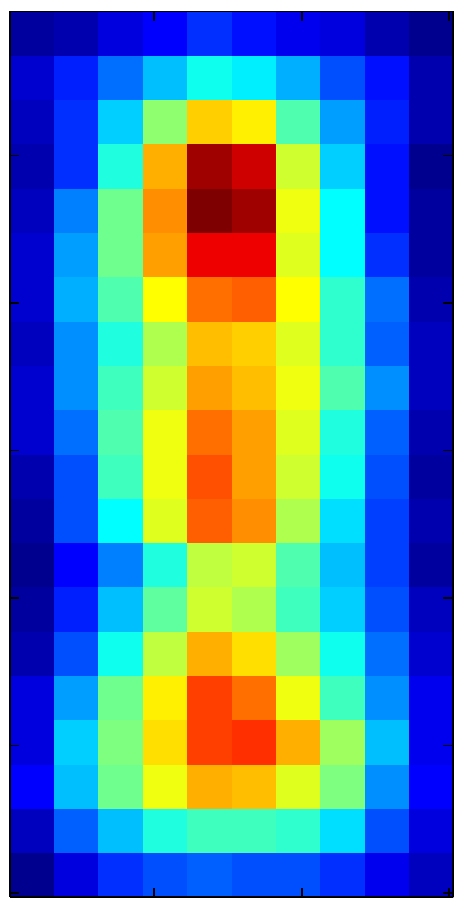}
\par\end{centering}

}
\par\end{centering}

\begin{centering}
\subfloat[\label{fig:supp-RandomFilters-model}Final \texttt{RandomFilters}
model]{\begin{centering}
\includegraphics[height=8em]{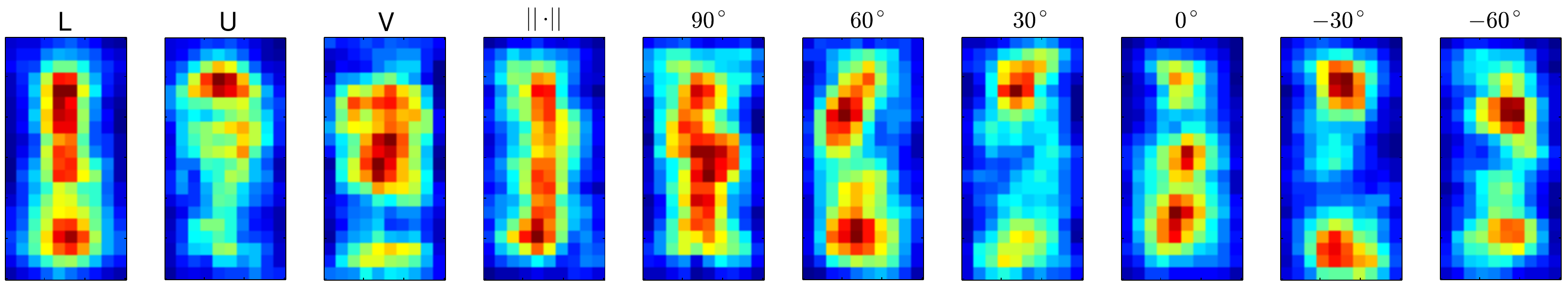}\qquad{}\includegraphics[height=8em]{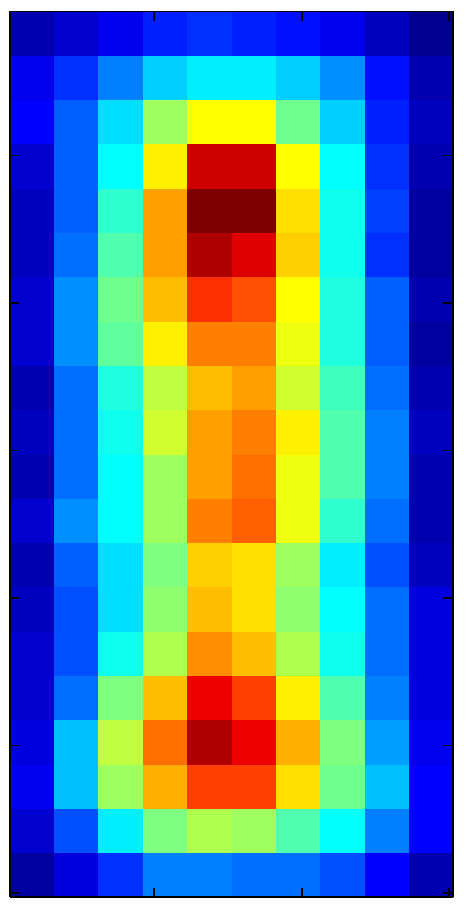}
\par\end{centering}

}
\par\end{centering}

\protect\caption{\label{fig:Spatial-distributions}Spatial distribution of learned
models. Per channel on the left, and across channels on the right.
Red areas indicate pixels that influence most the decision (used by
more decision trees). Figures \ref{fig:supp-Checkerboards4x3-model}
and \ref{fig:supp-RandomFilters-model} show our learned models (reach
$\sim\!\!18\%\ \mbox{MR}$ on Caltech test set), figure \ref{fig:supp-Filters-from-Roerei}
show a similar visualization for a weaker model ($\sim\!\!46\%\ \mbox{MR}$).
See text for discussion.}
\vspace{0em}
\end{figure*}

\begin{figure*}
\begin{centering}
\subfloat[Filters used in our final \texttt{Checker\-boards4x3 }model]{\begin{centering}
\includegraphics[width=0.8\textwidth]{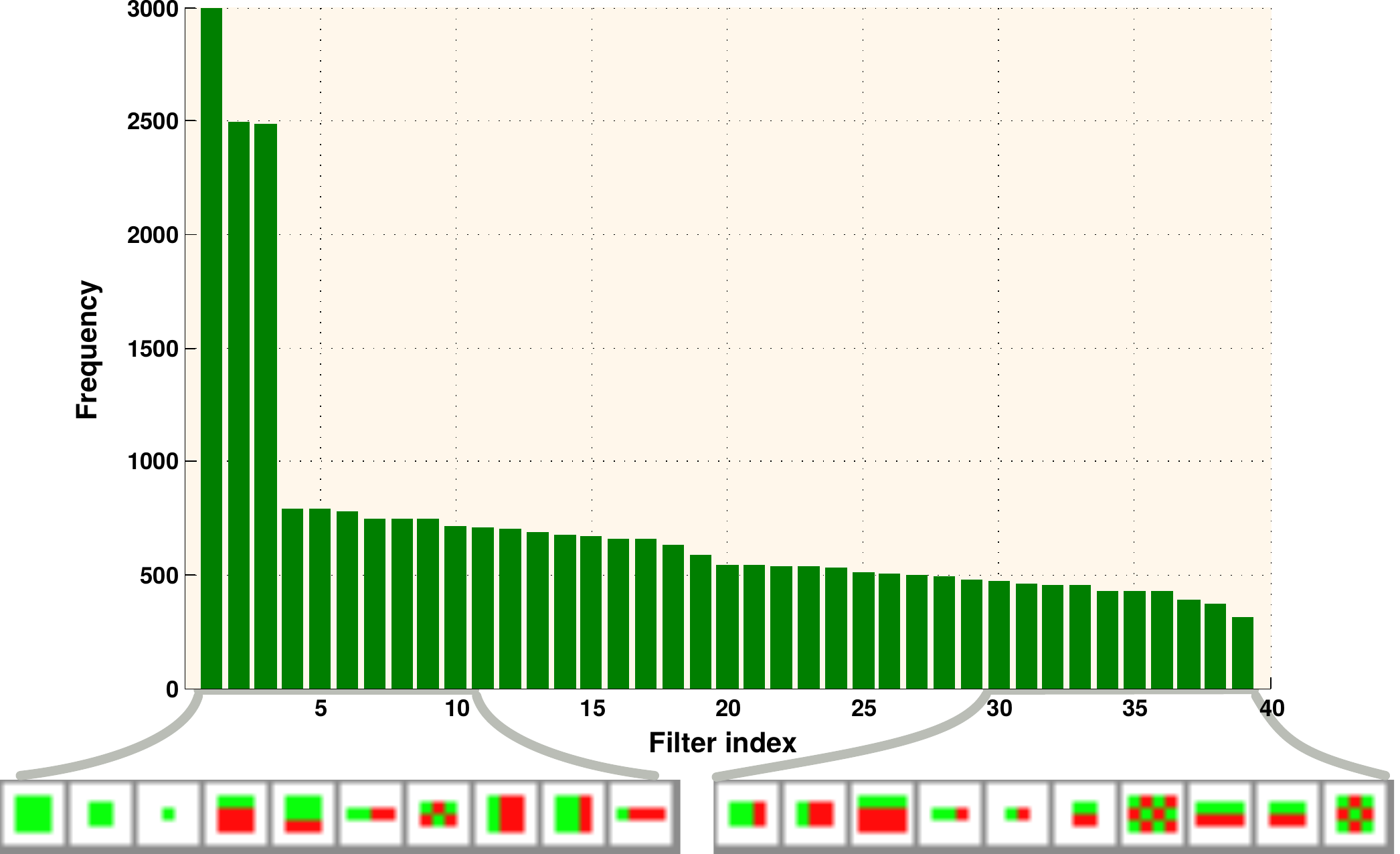}
\par\end{centering}

}
\par\end{centering}

\begin{centering}
\subfloat[Filters used in our final \texttt{RandomFilters} model]{\begin{centering}
\includegraphics[width=0.8\textwidth]{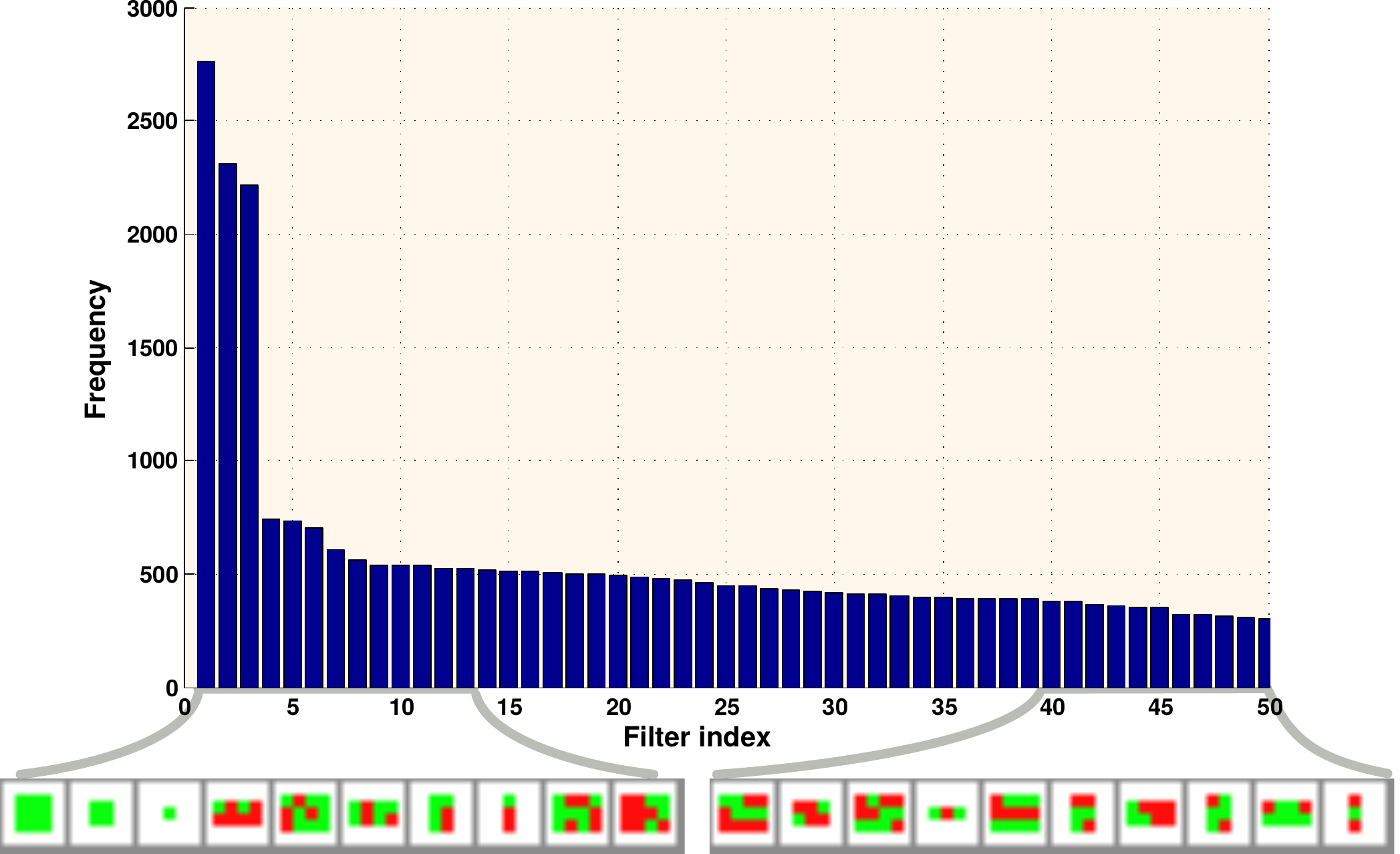}
\par\end{centering}

}
\par\end{centering}

\protect\caption{\label{fig:supp-Filters-frequency}Frequency of usage of each filter
as feature for decision tree split node (independent of the feature
channel). Left and right we show the top-10 and bottom-10 most frequent
filters respectively.\protect \\
Uniform filters are clearly the most frequently used ones (also used
in methods such as (\texttt{Roerei}, \texttt{ACF} and \texttt{(Squares)ChnFtrs}),
there is no obvious ordering pattern in the remaining ones. Please
note that each decision tree will probably use multiple filters across
multiple channels to reach its weak decision.}
\end{figure*}

In figures \ref{fig:Spatial-distributions} and \ref{fig:supp-Filters-frequency}
we present some qualitative aspects of the final learned models \texttt{Checker\-boards4x3}
and \texttt{RandomFilters} (see results section of main paper), not
included in the main submission due to space limitations.

In figure \ref{fig:Spatial-distributions} we compare the spatial
distribution of our models versus a significantly weaker model (\texttt{Roerei},
trained on INRIA, see figure 5 of main paper). We observe that our
strong models focalize in similar areas than the weak \texttt{Roerei}
model. This indicates that using filtered channels does not change
which areas of the pedestrian are informative, but rather that at
the same locations filtered channels are able to extract more discriminative
information.

In all three models we observe that diagonal oriented channels focus
on left and right shoulders. The \texttt{U} colour channel is mainly
used around the face, while \texttt{L} (luminance) and gradient magnitude
($\left\Vert \cdot\right\Vert $) channels are used all over the body.
Overall head, feet, and upper torso areas provide most clues for detection.

In figure \ref{fig:supp-Filters-frequency} we observe that the filters
usage distribution is similar across different filter bank families.

\end{document}